\documentclass[10pt,twocolumn,letterpaper]{article}

\usepackage[pagenumbers]{wacv} 

\usepackage{booktabs}
\usepackage{multirow}
\usepackage{graphicx}
\usepackage{array}

\definecolor{wacvblue}{rgb}{0.21,0.49,0.74}
\usepackage[pagebackref,breaklinks,colorlinks,allcolors=wacvblue]{hyperref}

\def\wacvPaperID{2172} 
\def\confName{WACV}
\def\confYear{2026}

\title{Weakly Supervised Ephemeral Gully Detection In Remote Sensing Images Using Vision Language Models}

\author{Seyed Mohamad Ali Tousi\textsuperscript{\textdagger}, Ramy Farag\textsuperscript{\textdagger},
John A. Lory\textsuperscript{\textdaggerdbl}, G. N. DeSouza\textsuperscript{\textdagger}\\
\small \textsuperscript{\textdagger}Vision Guided and Intelligent Robotics Laboratory (ViGIR), EECS Dept.\\
\small \textsuperscript{\textdaggerdbl}Division of Plant Science and Technology \\
\small University of Missouri
Columbia - MO - US\\
{\tt\small stousi, rmf3mc, loryj, desouzag@missouri.edu}}

\begin{document}
\maketitle
\begin{abstract}

Among soil erosion problems, Ephemeral Gullies are one of the most concerning phenomena occurring in agricultural fields. Their short temporal cycles increase the difficulty in automatically detecting them using classical computer vision approaches and remote sensing. Also, due to scarcity of and the difficulty in producing accurate labeled data, automatic detection of ephemeral gullies using Machine Learning is limited to zero-shot approaches which are hard to implement. To overcome these challenges, we present the \textbf{first weakly supervised pipeline for detection of ephemeral gullies. Our method relies on remote sensing and uses Vision Language Models} (VLMs) to drastically reduce the labor-intensive task of manual labeling. In order to achieve that, the method exploits: 1) the knowledge embedded in the VLM's pretraining; 2) a teacher-student model where the teacher learns from noisy labels coming from the VLMs, and the student learns by weak supervision using teacher-generate labels and a noise-aware loss function. We also make available the first-of-its-kind dataset for semi-supervised detection of ephemeral gully from remote-sensed images. The dataset consists of a number of locations labeled by a group of soil and plant scientists, as well as a large number of unlabeled locations. The dataset represent more than 18,000 high-resolution remote-sensing images obtained over the course of 13 years. Our experimental results demonstrate the validity of our approach by showing superior performances compared to VLMs and the label model itself when using weak supervision to train an student model. The code and dataset for this work are made  publicly available.

\end{abstract}
    
\section{Introduction} \label{Introduction}

Ephemeral Gullies (EG) have been designated as prominent causes of soil erosion \cite{soil}. Rain storms cause them to form in agricultural fields, and a lack of timely and accurate detection and treatment may lead to eventual soil and profit losses. The ephemeral nature of EGs makes it extremely hard to employ simple classical computer vision approaches in the process of their detection as they tend to appear and disappear through time. However, a temporal analysis of any given agronomic field can yield beneficial insights about their existence inside that same field. 

Despite the importance of the task, to the best of our knowledge the only labeled dataset for EG detection available in the literature is the benchmarking labeled data published in \cite{tousi2025zero}. However, this dataset does not provide enough data samples to train a deep learning model. This lack of labeled data is likely to be related to scarcity of supervised/semi-supervised approaches for EG detection, not to mention that most of the efforts in the area had to focus on zero-shot learning approaches, which can be much  harder to implement.

\begin{figure}[t!]
    \centering
    \includegraphics[width=0.99\linewidth]{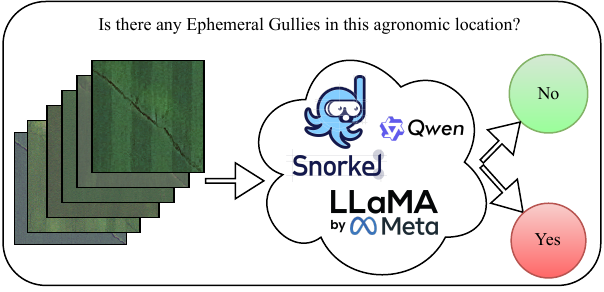}
    \caption{We propose to use a combination of Vision Language Models (VLMs) and Weak Supervision Frameworks (WSF) to train a classifier that detects the presence of Ephemeral Gullies (EGs) in agricultural fields.}
    \label{fig:intro}
\end{figure}

Recently, VLMs have shown promising performances in problems requiring zero-shot detection \cite{toubal2024modeling}. In a recent work \cite{tousi2025zero}, the authors incorporated the knowledge embedded in VLM pretraining into the task of EG detection and developed three zero-shot detection pipelines consisting of Visual Question Answering (VQA) and Large Language Models (LLMs). This development made possible to approach the problem of EG detection without requiring labor-intensive human labeling processes. Those pipelines paved a path to produce ``some" labels for large-scale EG detection datasets, however noisy and weak. An accurate and research-backed way of aggregating those noisy labels could make possible the training of a deep learning model on a large-scale EG detection dataset in a supervised manner.
In that regard, \cite{tousi2025combining} showed that using a Weak Supervision Framework (WSF) combined with VLMs can potentially lead to better detection performances in fine-grained classification tasks. A WSF estimates the accuracy of each ``weak" labelers (in this case the VLMs) and aggregates their labels accordingly. An end classifier model can then be trained on the pseudo-labels produced by the WSF, using a \textit{noise-aware} loss function \cite{ratner2017snorkel}. 

In this work, we propose to use a combination of weak supervision and multiple VLM-based pipelines proposed in \cite{tousi2025zero} to train an EG detection classifier network which outperforms the previous zero-shot methods. To summarize, our contributions are as follows:

\begin{enumerate}
    \item We incorporate a WSF into the EG detection process, enabling the training of deep learning models in a supervised manner while outperforming previous zero-shot methods. The proposed model is more accurate, even though it is lighter in terms of size when compared to the VLMs employed in zero-shot approaches. This makes the implementation of the trained network more practical in real-world EG detection scenarios.

    \item We propose a major improvement over the EG detection dataset published in \cite{tousi2025zero}. First, we increased the temporal information available -- i.e. from 6 to 8 remotely sensed images per location. Second, we included higher resolution images, providing a combination of 1m and 15cm Ground Sample Distances (GSDs) into the dataset. Third, we divided the data into two subsets: 1) a smaller labeled portion which was labeled by a group of soil and plant scientists, and 2) a large-scale unlabeled dataset (around 18,000 locations) to be used in semi-supervised/unsupervised training scenarios.  
    
    \item We propose the use FlexiViT \cite{beyer2023flexivit} as the feature extraction backend.  FlexiViT incorporates the power of Vision Transformers (ViT) in feature extraction while allowing the use of different patch sizes resulting from the combination of various GSDs in the dataset.

\end{enumerate}

\section{Related Works} \label{Related Works}

To this date, most of the research in the area of EGs are focused on determining the existence of EG by employing prior knowledge of their existence. In other words, instead of remote detection, manual assessment of the erosion features associated with EGs had to be well-researched and published in the literature so new predictions could be made manually \cite{Identifying, w12020603, china1, china2, reece2023using}. 


In the meantime, Visual Instruction Tuning was introduced by \cite{liu2023visual, liu2024improved} through LLaVA models (Large Language and Vision Assistant) to generate a large-scale visual assistant model. These models have been employed in zero-shot scenarios, more specifically in fine-grained classification tasks \cite{tousi2025combining}. Instruction Tuning was then used in developing state-of-the-art Llama3.2 Models \cite{meta2024llama}, which outperformed many open and closed-source vision models on common industry benchmarks. In light of those developments, the authors of \cite{tousi2025zero} have recently proposed the first pipelines for the detection of EGs using remote-sensed images. In other words, given that VLMs have shown promising results in fine-grained zero-shot classification tasks, they approached the scarcity of labeled data in EG detection using LLMs, VLMs and VQAs in a zero-shot manner to detect the presence of EGs within an agricultural field while they defined the problem as a binary classification task. This work was built on their previous work in \cite{toubal2024modeling}, and it proposed a VQA-based detection pipeline to help Llama3.2-Vision to identify the presence of an EG within an agronomic field more accurately. At the same time, they also demonstrated promising results in direct EG detection without incorporating VQA using a pipeline based purely on Qwen2.5-VL \cite{bai2023qwen}.
However, their individual zero-shot performance in EG detection has proven not to be sufficiently accurate and despite the novelty and importance of their work, implementing those large VLMs in real-world detection scenarios remains impractical.

\section{Method} \label{Method}

Before presenting the proposed pipeline to detect EGs using WSF and VLMs, a mathematical definition of the problem is necessary. After defining the problem, we present the proposed pipeline and its components in detail. 

\begin{figure*}
    \centering
    \includegraphics[width=0.99\linewidth]{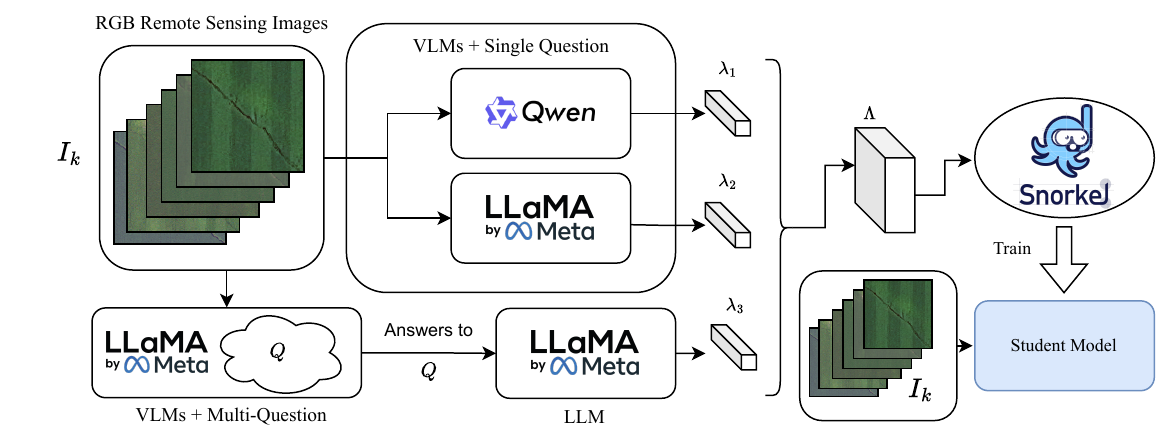}
    \caption{The proposed EG detection pipeline. The remote sensing RGB images are fed to the VLMs (Qwen2.5-VL and Llama3.2-Vision \cite{bai2023qwen, meta2024llama}) with two different prompting paradigms: 1) single question, and 2) multi-questions. The resulted noisy labels produced by the VLMs are being used to train a probabilistic label model (Snorkel \cite{ratner2017snorkel}). The trained label model then produce pseudo-labels which will be used in training an student model.}
    \label{fig:pipeline}
\end{figure*}

\subsection{Problem Definition}

As in \cite{tousi2025zero}, we define the EG detection problem as a binary classification task in which the classifier acts based on the presence of an EG in a given agricultural region. Let's assume that a dataset $ \mathcal{D}_u = \{ I_k | k = 1,...,K  \}$ exists for total $K$ agricultural locations in which each $ I_k = \{ i_n| n=1,...,N\}$ represents a set of $N$ remote sensing images taken over a temporal period of time. Now, a classifier $f_\theta$, in which $\theta$ is the set of tunable parameters, is desired to take each set of $I_k$ and produce a classification based on presence of an EG in the $k_{th}$ region of the dataset. In other words:

\begin{equation}
    c_k = f_\theta(I_k = \{i_1, i_2, ..., i_N\})
\end{equation}
Where $c_k$ is the predicted label for $k_{th}$ agronomic location. 

For evaluating any given classifier $f_\theta$, a labeled dataset $ \mathcal{D}_l = \{ (I_j,l_j) | j = 1,...,J  \}$ in required where $l_j$ is the ground-truth label for the $j_{th}$ agronomic location.

\subsection{Proposed Pipeline}

Our WS+VLM-based classification pipeline consists of three major components (which are depicted graphically in Figure \ref{fig:pipeline}): 

\begin{enumerate}
    \item Employing VLMs as noisy labeling functions.
    \item Optimizing a WS-based label model based on the labels produced by the VLMs.
    \item Training an student model with \textit{noise-aware} loss function on the pseudo labels coming from the optimized label model. 
\end{enumerate}

\subsubsection{VLMs as Labeling Functions}

Two different prompting paradigms are employed in the process of using VLMs to label each $I_k$:

\begin{enumerate}
    \item \textbf{Single-Question Pipeline:} In a single question pipeline, the VLMs are prompted to answer to a single question regarding the presence of an EG in the given set if images associated with $I_k$. In other words:

    \begin{equation}
        \lambda^{sq}_k = VLM(I_k,q)
    \end{equation}
    Where $\lambda^{sq}_k$ is the predicted label by the VLM in a single question paradigm and $q$ is the questioning prompt. 

    \item \textbf{Multi-Question Pipeline:} To prevent hallucinations by some of the VLMs and inspired by \cite{toubal2024modeling, tousi2025zero}, a set of different questions were used to prompt the VLMs that each of them targets one single visual attributes corresponding to presence of an EG. An LLM aggregates the answers to those questions and produces the final classification. In other words:

    \begin{equation}
        \lambda^{mq}_k = LLM(VLM(I_k,Q),p)
    \end{equation}
    Where $\lambda^{mq}_k$ is the predicted label by the VLM in a multi-question paradigm, $Q$ is the set of questions, and $p$ is the prompt for LLM to aggregate the responses coming from the VLM. 
\end{enumerate}

Both of the above-mentioned prompting paradigms are depicted in Figure \ref{fig:pipeline}. 

\subsubsection{Weak Supervision Label Model}

Assuming $m$ labeling functions were used in the previous step, an array of predicted labels is developed as follows:

\begin{equation}
    \Lambda = \begin{bmatrix}
\lambda_{1,1} & \cdots & \lambda_{1,m} \\
\vdots & \ddots & \vdots \\
\lambda_{K,1} & \cdots & \lambda_{K,m}
\end{bmatrix}
\end{equation}
Where $K$ is the number of locations in the dataset and $m$ is the number of labelers. 

Now, a WSF (in this case Snorkel \cite{ratner2017snorkel}) is employed to estimate the accuracy of each of the labelers. 

Snorkel employs a generative label model \( p_{w}(\Lambda, Y) \), defined in Equation~(\ref{eq:p_w}), to aggregate the \textit{weak} labels provided by the VLMs:

\begin{align}\label{eq:p_w}
    p_{w}(\Lambda, Y) = Z_{w}^{-1} \exp\Bigg( 
\sum_{i=1}^{m} w^{T} \phi_{i}(\Lambda, y_{i}) \Bigg)
\end{align}

Here, \( Z_{w} \) is a normalization constant, \( w \) is a vector of model parameters (weights), and \( \Lambda \) is the array of noisy labels obtained in the previous step. The true labels are denoted by \( Y \), and the feature vector \( \phi_i \) is a concatenation of the following three types of factors:

\[
\begin{aligned}
\phi^{\text{Lab}}_{k,m}(\Lambda, Y) &= \mathbf{1}\{\Lambda_{k,m} \neq \emptyset\} \\
\phi^{\text{Acc}}_{k,m}(\Lambda, Y) &= \mathbf{1}\{\Lambda_{k,m} = y_k\} \\
\phi^{\text{Corr}}_{k,j,d}(\Lambda, Y) &= \mathbf{1}\{\Lambda_{k,j} = \Lambda_{k,d}\} \quad \text{for } (j,d) \in C
\end{aligned}
\]
Where \( C \) is the set of potential correlations between the labeling functions. 
Since the true labels \( Y \) are not available, the model is trained by minimizing the negative log marginal likelihood over the observed label matrix \( \Lambda \):

\[
\hat{w} = \arg\min_{w} \left( -\log \sum_{Y} p_w(\Lambda, Y) \right) .
\]

Once the label model is trained on \( \Lambda \), it is ready to produce probabilistic pseudo-labels for training an student classifier.

\subsubsection{Student Model}

A student model to be trained on EG detection task requires to have two main characteristics:

\begin{enumerate}
    \item A powerful feature extraction backend suitable for remote sensing images. 

    \item The ability to process different resolution images (when dealing with remote sensing images with different Ground Sample Distances (GSD), resizing images often is not a suitable option). 
\end{enumerate}

To that end, we use FlexiViT \cite{beyer2023flexivit} as our feature extraction backend. FlexiViT incorporates the power of Vision Transformers (ViT) in feature extraction while allowing the use of different patch sizes. The ability to change the patch size allows us to maintain the original image resolution and process each image in $I_k$ without losing any important information in the process of resizing. 

\begin{figure}
    \centering
    \includegraphics[width=0.99\linewidth]{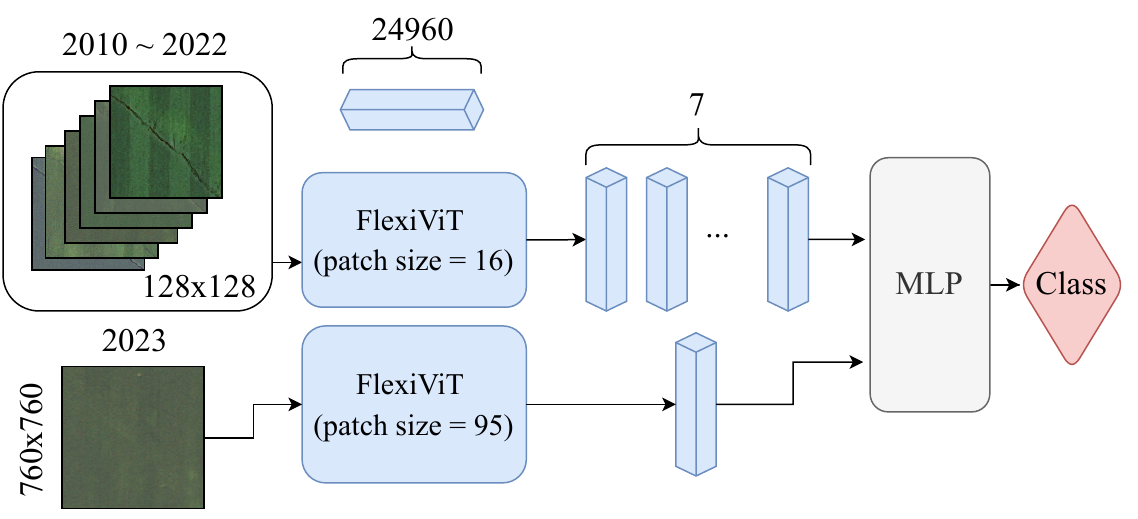}
    \caption{The proposed student model uses FlexiViT \cite{beyer2023flexivit} as its feature extractor backbone. The patch sizes are chosen specifically to represent the same geographical area in both low and high resolution images. An MLP aggregates the extracted features and provides the final classification.}
    \label{fig:flexivit-pipeline}
\end{figure}

Figure \ref{fig:flexivit-pipeline} shows the schematic of the student model. Each image in \( I_k\) is fed to the FlexiViT with suitable patch size for its features to be extracted. \textbf{The patch sizes are chosen specifically to represent the same geographical location across the images}. After that, a Multi-Layer Perceptron (MLP) takes the extracted features across all the images from \( I_k\) and produces a classification. 

The objective here is to train a discriminative model (classifier) that generalizes beyond the weak supervision labels provided by the labeling functions. In our case, the MLP is denoted by \( f_{\theta} \), and we train it on the probabilistic labels \( \tilde{Y} \) inferred by the generative model. To account for label uncertainty, we minimize a \textit{noise-aware} version of the loss function \( \ell(f_{\theta}(I_k), y) \), which is the expected loss with respect to the probabilistic label distribution \( \tilde{Y} \):

\[
\hat{\theta} = \arg\min_{\theta} \sum_{i=1}^{m} \mathbb{E}_{y \sim \tilde{Y}} \left[ \ell(f_{\theta}(I_k), y) \right] .
\]

\section{EG Detection Dataset} \label{Dataset}

 We have developed a labeled EG detection evaluation dataset as well as a large scale unlabeled dataset for semi-supervised/unsupervised EG detection approaches. We first present the detailed specification of the dataset and then provide information about the labeling process. 

\subsection{Dataset Specifications}

Our dataset consists of 128mx128m areas of agronomic locations in which each location has 
eight remotely sensed RGB images taken temporally ranging from years 2010 to 2023. Two major sources of RGB images were used in the dataset: 1) USDA National Agriculture Imagery Program (NAIP) \cite{naip_usda} for lower resolution images (100cm GSD), and 2) Missouri Spatial Data Information Service (MISDIS) \cite{msdis_dataset} for high resolution images (15cm GSD). 

\subsection{Unlabeled Dataset}

For the large-scale unlabeled dataset for EG detection, we generated over 18,000 locations (more than 144,000 individual images) inside three subwatersheds (HUC12) in the state of Missouri. Each subwatershed has an area of approximately 40 to 160 squared kilometers. The VLMs were used to label each location in terms of EG presence in the region.

\subsection{Labeled Evaluation Dataset}

In order to generate our labeled evaluation dataset, we relied on the database from \cite{reece2023using} in which \textbf{regions with EGs} were delineated in numerous HUC12s using Google Earth historical images inside four states of Missouri, Kansas, Iowa and Nebraska. We chose three HUC12s in Missouri (for which our MISDIS data is available) and generated positive and negative examples with respect to the regions delineated in their database. Around 500 locations were generated for the evaluation set that combines positive and negative samples. However, because of the change in imagery and their time of capture (from Google Earth to NAIP and MISDIS), and also the ephemeral nature of EGs, the initial labels (coming from the \cite{reece2023using}) are not sufficiently accurate for our dataset. Hence, a separate labeling process by soil and plant scientists has been performed to ensure that high-quality ground truth labels are available for our evaluations. The details of the labeling process are presented in section \ref{dataset-generation}.

Both labeled and unlabeled datasets are available for download through this link: \url{http://vigir.ee.missouri.edu/Research/GullyDetection/}.

\section{Experiments and Results}\label{Experiments}

To showcase the validity of the proposed approach to detect EGs, we answered these research questions:

\begin{enumerate}

    \item How good is the performance of each individual VLM in detecting the presence of EGs?

    \item How well does the WSF label model performs in EG detection?

    \item Finally, how well does the student model trained with \textit{noise-aware} loss function on WSF-generated pseudo labels perform in EG detection task?
    
\end{enumerate}

Before answering these questions, we present some implementation details about our proposed pipeline. 

\subsection{Implementation Details}

Table \ref{tab:impl_details_grouped} shows the implemented parameters and hyperparameters in the proposed pipeline. 

\begin{table}[ht]
\centering
\footnotesize
\begin{tabular}{@{}>{\raggedright}p{2cm} p{3cm} p{3cm}@{}}
\toprule
& \textbf{Component} & \textbf{Value / Choice} \\
\midrule

\multirow{2}{*}{\rotatebox[origin=c]{0}{\textbf{VLM Prompting}}}
& Framework & Ollama \cite{ollama2024github} \\
& Hardware Used & 4 x NVIDIA A100 \\

\midrule

\multirow{2}{*}{\rotatebox[origin=c]{0}{\textbf{Label Model}}}
& Framework & Snorkel \cite{ratner2017snorkel} \\
& Number of Epochs & 100 \\

\midrule

\multirow{4}{*}{\rotatebox[origin=c]{0}{\textbf{Student Model}}}
& Framework & PyTorch 2.0 \\
& Loss Function & KL Divergence \\
& MLP & [8x24960, 2] \\
& Hardware & 8 x NVIDIA A10 \\

\bottomrule
\end{tabular}
\caption{Implementation details grouped by model design, training setup, and system configuration.}
\label{tab:impl_details_grouped}
\end{table}

To implement the noise-aware loss function using the built-in losses in PyTorch, we optimize the marginalized KL divergence over the label distribution as shown in equation \ref{eq:kl}: 

\begin{equation}\label{eq:kl}
\footnotesize
\mathrm{KL}(\tilde y \,\|\, p_\theta)
= \sum_{c=1}^K \tilde y_c \log \frac{\tilde y_c}{p_\theta(c)}
= \underbrace{\sum_{c=1}^K \tilde y_c \log \tilde y_c}_{\text{constant}}
\;-\;\sum_{c=1}^K \tilde y_c \log p_\theta(c).
\end{equation}

where $K$ denotes the number of classes, $\tilde{y}$ represents the (potentially noisy) label distribution, $\tilde{y}_c$ is the probability mass assigned to class $c$ under $\tilde{y}$, and $p_\theta(c)$ is the predicted probability of class $c$ under the model with parameters $\theta$. Dropping the constant term is equivalent to minimizing the expected cross-entropy.

\subsection{Experimental Results}

Table \ref{tab:model_comparison} summarizes answers to all the three research questions mentioned above. First of all, the three first rows show the individual performance of the VLMs on the evaluation set. The results are in agreement with the findings in \cite{tousi2025zero}, which showed better language understanding in Llama3.2-Vision in multi-question scenario and better visual understanding in Qwen2.7-VL:72B in a single-question approach. We have also trained our student model with the database in \cite{reece2023using} in a fully supervised manner. Finally, the last two rows in Table \ref{tab:model_comparison}  show the performance of the label model itself on the evaluation set and the student model when trained on large scale dataset with weak supervision of the label model. 

\begin{table*}[ht]
\centering
\begin{tabular}{p{7cm}ccccc}
\toprule
\textbf{Model / Approach} & \textbf{NPV} & \textbf{Recall} & \textbf{Precision} & \textbf{F1 Score} & \textbf{Accuracy} \\
\midrule
Llama3.2-Vision:90B / Single Question & 0.55 & 0.42 &\textbf{0.73} & 0.53 & 0.61 \\
Llama3.2-Vision:90B / Multiple Question & 0.56 & 0.80 & 0.57 & 0.66 & 0.57 \\
Qwen2.7-VL:72B / Single Question & 0.70 & \textbf{0.83} & 0.63 & 0.72 & 0.65 \\
\midrule
SM + \cite{reece2023using} & 0.64 & 0.81 & 0.60 & 0.69 & 0.61 \\
\midrule
LM & 0.65 & 0.76 & 0.64 & 0.69 & 0.64 \\
\midrule
\textbf{SM + LM + Large Scale Dataset} 
& \textbf{0.74 (+4\%)} & 0.81 & 0.68 & \textbf{0.74 (+2\%)} & \textbf{0.70 (+5\%)} \\
\bottomrule
\end{tabular}
\caption{Performance comparison of various VLMs and supervision strategies on the evaluation set. SM and LM stand for student model and label model respectively. We are also reporting the Negative Predictive Value (NPV) alongside the other metrics to be used in practical evaluations.}
\label{tab:model_comparison}
\end{table*}

\section{Discussion} \label{Discussion}

\subsection{On VLM Choices and Prompting Paradigms}

The choice of VLMs used in this study was based on research in \cite{tousi2025zero}. In that study, it was shown that Qwen2.5-VL is the best model when dealing with single question prompting paradigm in detection EGs. Conversely, if used in multi-question scenarios, Qwen2.5-VL did not show a satisfactory performance. Llama3.2-Vision, on the other hand, excelled in multi-question paradigm. Table \ref{tab:model_comparison} shows that the F1 score of Llama3.2-Vision used in multi-question scenario was meaningfully higher than the single question scenario. 

The questions used in both single and multi-question prompting paradigms were exactly the same as ones proposed in \cite{tousi2025zero}. In single question scenario, one question targeting the presence of an EG in the given agronomic region was presented to the VLM alongside 8 temporal images from that region. In multi-question paradigm, the 15 best questions from \cite{tousi2025zero} were chosen and presented to the VLM.

\subsection{On Data Generation and Evaluation Ground Truth}\label{dataset-generation}

The large scale dataset for EG detection provided by this work was generated randomly but within boundaries of carefully chosen subwatersheds inside the state of Missouri. These subwatersheds were analyzed in terms of EG presence before (in \cite{reece2023using}). 


For the labeled evaluation dataset, random sampling the locations was not a suitable option as it was highly probable that by randomly generating an evaluation set, the dataset becomes extremely imbalanced toward negative samples (there are much more locations \textit{without} EGs than with them). Given the smaller size of the evaluation dataset, this problem would have been more emphasized. To tackle the problem of dataset imbalance a database provided by \cite{reece2023using} was used to generate a roughly balanced evaluation set of locations. However, a separate round of labeling process was necessary due to the difference between the images associated with each location and their respective time of the capture.

In the labeling process, the expert labelers were presented with 8 images for any given location. For each image they had the option of choosing 5 levels of confidence in presence of an EG: 
\begin{itemize}
    \item \textbf{0}: Absolutely certain - EG negative.

    \item \textbf{1}: Almost certain - EG negative.

    \item \textbf{2}: Not sure - Irrelevant Location - Insufficient information.

    \item \textbf{3}: Almost certain - EG positive.

    \item \textbf{0}: Absolutely certain - EG positive.
\end{itemize}

Four different voting schemes were adopted to aggregate the labels from soil and plant experts:

\begin{enumerate}
    \item \textbf{Strict Positive}: All the labelers should find at least one image with label 4 and one image with label bigger than 2, otherwise counted as negative. 

    \item \textbf{Lenient Positive}: All the labelers should find at least one image with label 4, otherwise counted as negative. 

    \item \textbf{Lenient Negative}: All the labelers should not find any image with a label greater than 1, otherwise counted as positive.

    \item \textbf{Strict Negative}: All the labelers should not find any image with a label greater than 0, otherwise counted as positive.

\end{enumerate}

 Table \ref{tab:gt_metrics} shows the comparison between the results of our four different labeling schemes with the initial labels coming from the database in \cite{reece2023using}.

\begin{table}[ht]
\centering
\footnotesize
\begin{tabular}{lcccc}
\midrule
\textbf{GT/Metric} & \textbf{Accuracy} & \textbf{Recall} & \textbf{Precision} & \textbf{F1 Score} \\
\midrule
Strict Positive     & 0.521 & 0.592 & 0.690 & 0.523 \\
Lenient Positive    & 0.528 & 0.598 & 0.694 & 0.532 \\
Lenient Negative    & 0.671 & 0.686 & 0.712 & 0.683 \\
\textbf{Strict Negative}   & \textbf{0.697} & \textbf{0.698} & \textbf{0.727} & \textbf{0.706} \\
\midrule
\end{tabular}
\caption{Metrics of comparison with database in \cite{reece2023using}, under different ground truth  generation schemes.}
\label{tab:gt_metrics}
\end{table}

As the evaluation locations were generated with respect to the regions delineated in \cite{reece2023using}, a better agreement with their database (as seen our strict negative labeling scheme) was preferred. So, the strict negative scheme were chosen to produce the final ground truth labels. Another reason for choosing the fourth voting scheme was that in practical scenarios, false negatives are posing more problem than false positives. This is because neglecting the presence of an EG in an agronomic field can cause more agronomic problems. So, having a ground truth label that is strict when labeling a location as negative is more favorable. 

\subsection{On WSF and Label Model Performance}

As table \ref{tab:model_comparison} shows, the performance of the label model was slightly lower than that of the best VLM. This result was expected, as the label model produced probabilistic labels for which all the labelers contributed. Binarization of those probabilities and their comparison with the ground truth reduces the label model to voting scheme between the labelers. The performance of the label model itself can be improved by 1) incorporating more labeling functions to the pipeline, and 2) having better labeling functions with more agreements \cite{tousi2025combining}.

\subsection{On Student Model Results}

Not surprisingly, when we trained the student model with the labels extracted from \cite{reece2023using}, the performance was not as high. This was another reason for us to develop a separate set of ground truth labels for the newly generated evaluation set. If the labels from \cite{reece2023using} were perfect, a model trained on those labels in a fully-supervised manner should have performed much better than any given model trained with weak supervision or using an unsupervised approach. However, table \ref{tab:model_comparison} shows that training the student model with \cite{reece2023using} was not outperforming the best VLM and/or the weakly supervised model.

The promise of weak supervision was to train an end classifier that generalizes beyond the information embedded in the labeling functions. The results in table \ref{tab:model_comparison} show that this promise was met, and the student model trained with the weak supervision of the label model had a clear advantage (more than 5\% improvement in accuracy) over all the labeling functions (VLMs) and the label model itself. Other cases in the same table my have outpeformed the proposed method in Recall/Precision, but at the high price of low Precision/Recall. Improvement in NPV was also quite promising for real-world EG monitoring scenarios, in which false negative predictions pose higher risks than false positive ones.

\section{Conclusion} \label{Conclusion}

We presented the first weak supervision pipeline to train an EG detector using VLMs as weak labeling functions. The VLMs, prompted with two different paradigms, produced weak labels for a large scale remote sensing dataset regarding the presence of an EG in each location. Then a WSF aggregated the weak labels and provided probabilistic pseudo-labels to train an end classifier with noise-aware loss function. To evaluate the results, we developed and made available the first public semi-supervised dataset for EG detection that consists of a large scale unlabeled dataset with more than 18,000 locations, and a smaller labeled evaluation set. The ground truths for the evaluation set were produced by multiple soil and plant scientists. Our experimental results showed that the end classifier model, trained with the weak supervision of the WSF, is capable of generalizing beyond the information provided by the labeling functions and can improve the performance of the VLMs and the label model itself by a meaningful margin.  
{
    \small
    \bibliographystyle{ieeenat_fullname}
    \bibliography{main}
}

\end{document}